\documentclass{article}

\usepackage[english]{babel}
\usepackage{float}
\usepackage[letterpaper,top=2cm,bottom=2cm,left=3cm,right=3cm,marginparwidth=1.75cm]{geometry}
\usepackage{multirow}
\usepackage{amsmath}
\usepackage{graphicx}
\usepackage[colorlinks=true, allcolors=blue]{hyperref}
\usepackage{cite}
\usepackage{booktabs}

\title{NovisVQ: A Streaming Convolutional Neural Network for No-Reference Opinion-Unaware Frame Quality Assessment}

\author{
\begin{tabular}{cc}
Kylie Cancilla & Alexander Moore \\
\texttt{cancilla5@llnl.gov} & \texttt{moore278@llnl.gov} \\[1em]
Amar Saini & Carmen Carrano \\
\texttt{saini5@llnl.gov} & \texttt{carrano2@llnl.gov}
\end{tabular}
}

\begin{document}

\maketitle
\begin{abstract}
 Video quality assessment (VQA) is vital for computer vision tasks, but existing approaches face major limitations: full-reference (FR) metrics require clean reference videos, and most no-reference (NR) models depend on training on costly human opinion labels. Moreover, most opinion-unaware NR methods are image-based, ignoring temporal context critical for video object detection. In this work, we present a scalable, streaming-based VQA model that is both no-reference and opinion-unaware. Our model leverages synthetic degradations of the DAVIS dataset, training a temporal-aware convolutional architecture to predict FR metrics (LPIPS , PSNR, SSIM) directly from degraded video, without references at inference. We show that our streaming approach outperforms our own image-based baseline by generalizing across diverse degradations, underscoring the value of temporal modeling for scalable VQA in real-world vision systems. Additionally, we demonstrate that our model achieves higher correlation with full-reference metrics compared to BRISQUE, a widely-used opinion-aware image quality assessment baseline, validating the effectiveness of our temporal, opinion-unaware approach.
\end{abstract}

\section{Introduction}
Video is increasingly central to applications ranging from mobile devices and surveillance to autonomous laboratories and self-driving cars. In these domains, video object detection plays a critical role, yet its reliability is highly sensitive to real-world degradations such as blur, compression artifacts, and network-induced distortions. Accurate video quality assessment (VQA) is therefore essential for maintaining robust performance in detection pipelines.

Conventional VQA approaches are limited. Full-reference (FR) metrics such as LPIPS \cite{zhang2018unreasonable}, PSNR and SSIM \cite{wang2004image} provide objective scores but require clean reference videos for comparison, which are rarely available. No-reference (NR) methods remove this dependency but often rely on expensive, subjective human opinion scores for training. Furthermore, most existing NR models are image-based, overlooking the temporal continuity that is especially relevant for video object detection.

We address these challenges with a novel, streaming-based, no-reference, opinion-unaware VQA model. Our method uses synthetic degradations of the DAVIS dataset \cite{caelles2019davis} and self-supervised labels from FR metrics (LPIPS, PSNR, SSIM) to train a convolutional architecture that predicts per frame video quality directly from degraded video. By leveraging temporal context, our model greatly improves upon an image-based baseline. We demonstrate that our approach provides accurate quality estimates.

The remainder of this paper is organized as follows:  Section \ref{sec:contributions} details the key contributions of our research. Section \ref{sec:related_work} provides an overview of related work in video quality assessment. Section \ref{sec:Methods} describes our proposed model architecture and training methodology. Finally, Section \ref{sec:Results} presents our experimental results and analysis, followed by a conclusion in Section \ref{sec:Conclusion}.

\subsection{Contributions}
\label{sec:contributions}
Our primary contributions are as follows:
\begin{enumerate}
\item We demonstrate that streaming architectures substantially outperform image-based approaches for video quality assessment, with temporal context enabling generalization to unseen degradations where image-based models fail. 
\item We propose NovisVQ, a no-reference, opinion-unaware model that predicts full-reference metrics (LPIPS, PSNR, SSIM) per frame using only degraded video input, trained via a scalable self-supervised methodology on synthetically augmented data.
\item We validate our approach on real-world motion blur, achieving strong correlation with ground truth despite training exclusively on synthetic degradations, demonstrating effective synthetic-to-real generalization.
\end{enumerate}

\subsection{Related Work}
\label{sec:related_work}
Video quality degradation significantly impacts computer vision systems, making video quality assessment (VQA) crucial for robust pipelines. VQA methods are categorized into full-reference (FR), reduced-reference (RR), opinion-aware no-reference (NR), and opinion-unaware no-reference approaches.

\textbf{Full-Reference (FR) Quality Metrics.} FR metrics like PSNR, SSIM, and LPIPS measure distortion against pristine references. LPIPS correlates well with human perception and downstream task performance \cite{steinhauser2025data}. Netflix's VMAF is widely used for Quality of Experience but requires references and human opinion scores, limiting scalability \cite{vmaf}. The reference requirement severely restricts FR metrics' practical applicability.

\textbf{Opinion-Aware No-Reference (NR) Quality Metrics.} State-of-the-art NR models \cite{guan2024qmamba, mittal2012norqa} train on large-scale human opinion datasets \cite{ponomarenko2009tid2008, hosu2020koniq, ghadiyaram2016massive}, such as LIVE-FB LSVQ with 5.5M image quality ratings \cite{ying2021patch}. While effective at predicting subjective quality, human dependency makes these models expensive, subjective, and difficult to scale for downstream task optimization. Moreover, these image-based approaches do not leverage temporal information available in video data.

\textbf{Opinion-Unaware NR-IQA/VQA.} Recent work explores objective, scalable alternatives \cite{venkatanath2015pique, ramesh2025hirqa}. Agnolucci et al. proposed text-based labels for image quality assessment without human perception data \cite{agnolucci2025quality}. Menon et al. developed a streaming model to predict FR metrics \cite{menon2023transcoding}, but their reduced-reference design assumes partial pristine video access, making it impractical for real-time pipelines without references.

Our approach develops a no-reference, opinion-unaware VQA model trained on objective FR metric labels from synthetically augmented videos. Unlike prior work, we demonstrate that leveraging temporal information through video-based modeling provides substantial benefits for robust, scalable quality assessment in computer vision pipelines.

\section{Methods}
\label{sec:Methods}

Our model, \textbf{NovisVQ}, estimates three full-reference quality metrics capturing diverse image quality measures (LPIPS, PSNR, and SSIM) using only degraded video input. We selected these three metrics because they capture complementary aspects of image quality: LPIPS measures perceptual similarity, PSNR quantifies pixel-level fidelity, and SSIM assesses structural information. The architecture combines a multiscale ResNet-based encoder, an LSTM module for temporal context, and a lightweight MLP that outputs the three predictions.

\subsection{Training Approach}

We first developed an image-based baseline, \textbf{NovisIQ}, by fine-tuning ConvNeXt Small \cite{liu2022convnet} on the DAVIS dataset \cite{caelles2019davis}. Both models were trained on synthetic distortions (blur, JPEG compression, brightness) and evaluated on unseen augmentations (noise, color jitter, saturation) to test generalization beyond memorization of augmentation-specific patterns.

Both models were optimized using mean absolute error (MAE) loss, averaged across the three target metrics:

\begin{equation}
    \text{Total Loss} = \frac{1}{3} \left( \text{MAE}_{\text{LPIPS}} + \text{MAE}_{\text{PSNR}} + \text{MAE}_{\text{SSIM}} \right),
\end{equation}

\noindent where $\text{MAE}_{m} = \frac{1}{N} \sum_{i=1}^{N} \left| \hat{y}_{i}^{(m)} - y_{i}^{(m)} \right|$ for $m \in \{\text{LPIPS}, \text{PSNR}, \text{SSIM}\}$.

While NovisIQ performed well on validation samples using the same augmentations that were used in training, it failed to generalize to unseen augmentations. NovisVQ's streaming design enabled it to leverage temporal context and learn relative frame quality, motivating our focus on streaming architectures for robust video quality assessment.

\subsection{Model Architecture}
\begin{figure}[H]  
    \centering
    \includegraphics[width=0.85\linewidth]{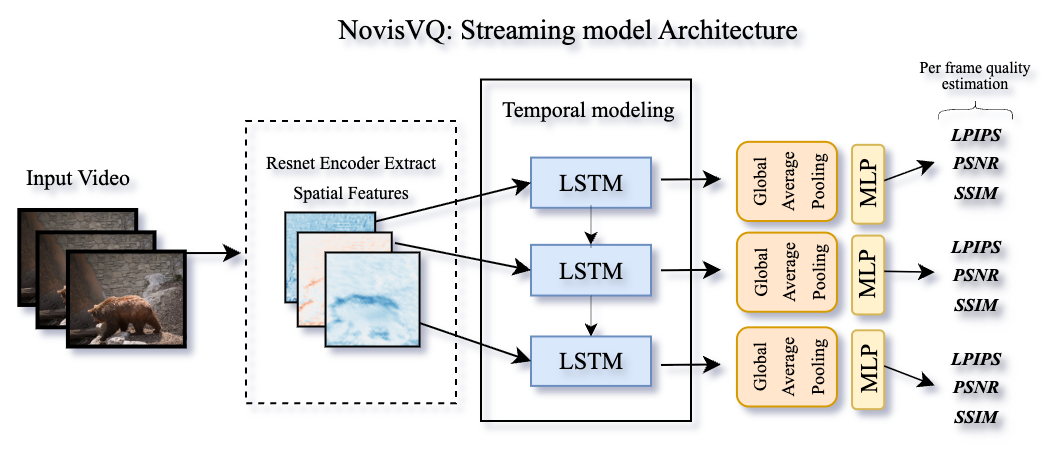}  
    \caption{NovisVQ architecture: multiscale ResNet encoder extracts features from degraded frames, LSTM captures temporal context, and MLP predicts LPIPS, PSNR, and SSIM scores.}
    \label{fig:architecture}
\end{figure}

\subsection{Quality Metrics and Augmentation Analysis}

We normalized all metrics to [0,1] to enable unified training, where 1 indicates high quality (augmented image has a low distance from ground truth) and 0 indicates poor quality (augmented image has a high distance from ground truth). These full-reference estimators measure quality by comparing augmented frames against pristine ground truth references. LPIPS computes perceptual similarity via patch activations and correlates well with human perception; we invert LPIPS scores (1 - LPIPS) so that higher values indicate better quality. PSNR measures pixel-wise fidelity as the ratio of maximum pixel value squared to MSE; we linearly normalize PSNR by clamping values to [10, 50] dB and mapping this range to [0,1], where higher values naturally indicate better pixel-wise accuracy. SSIM quantifies structural similarity, naturally bounded in [0,1] where higher values indicate better preservation of structural information.

\begin{figure}[H] 
    \centering
    \includegraphics[width=\linewidth]{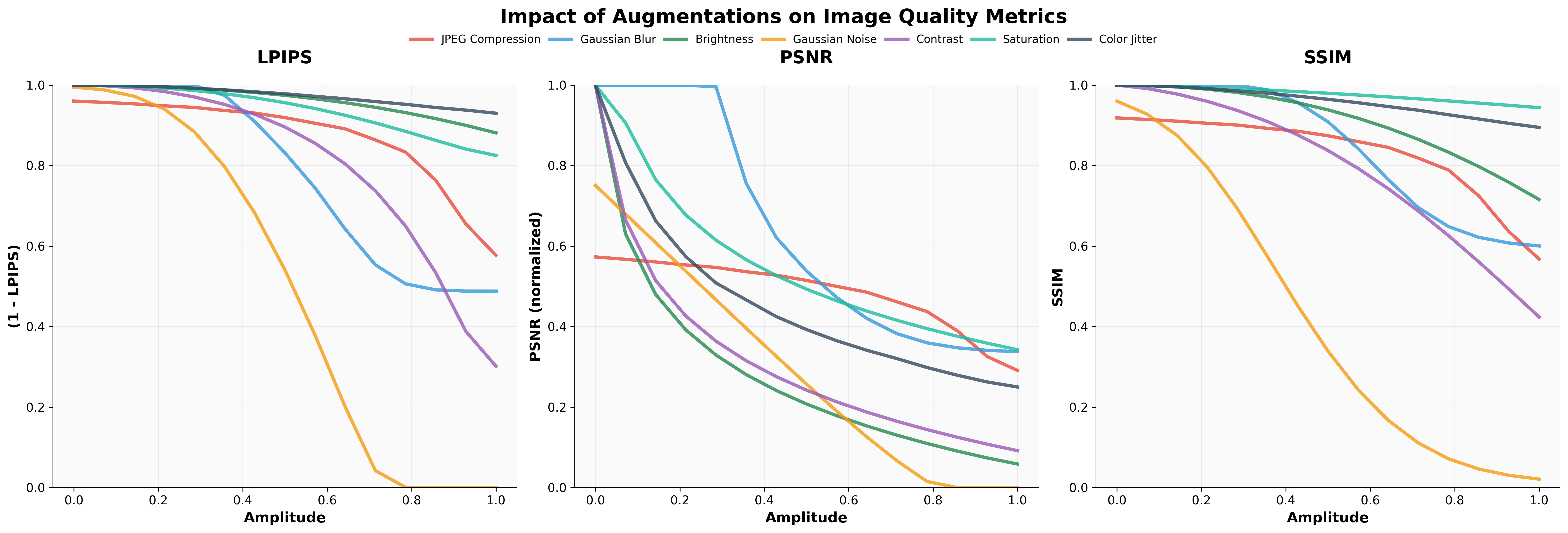}
    \caption{Impact of Synthetic Augmentations on LPIPS, PSNR and SSIM}
    \label{fig:augmentation_curves}
\end{figure}

Our analysis reveals distinct degradation profiles across augmentations. Gaussian noise causes near-complete quality collapse, with normalized metrics degrading by approximately 100\% for LPIPS and PSNR, and 97.8\% for SSIM relative to pristine video. In contrast, quality metrics show remarkable resilience to color-based augmentations: saturation retains 82.5\% of its initial LPIPS quality (a 17.5\% degradation) even at maximum amplitude. Notably, brightness and color jitter create severe PSNR degradation (94.1\% and 75.2\% decreases respectively) with minimal perceptual impact, indicating large pixel-level deviations while preserving semantic content-LPIPS degrades by only 8.4\% and 7.1\% for these augmentations. These distinct patterns motivated our train-test split: training on moderate-impact transformations (blur, brightness, JPEG) and testing on extreme (noise) or mild (color-based) variations. This augmentation-level split ensures the model must generalize to entirely unseen distortion types rather than memorizing augmentation-specific artifacts.

\section{Results}
\label{sec:Results}
\subsection{Loss Function and Validation Dynamics}

To evaluate generalization, we split videos by augmentation type: training augmentations (blur, brightness, JPEG compression) versus validation augmentations (Gaussian noise, saturation, color jitter). This tests whether models learn transferable quality features or simply memorize distortion-specific patterns.

NovisIQ quickly overfit to training distortions, with validation performance degrading across all metrics: total loss error increased 1.82 times, LPIPS prediction error 1.07 times, SSIM prediction error 1.20 times, and PSNR prediction error 3.67 times. Statistical analysis confirmed weak correlations with meaningful learning ($R^{2} < 0.01$, $p > 0.5$ for LPIPS and SSIM), indicating NovisIQ memorized augmentation-specific artifacts rather than learning transferable features.
NovisVQ continued improving throughout training and generalized strongly to unseen augmentations, reducing total loss error by 2.86 times, LPIPS prediction error by 5.07 times, SSIM prediction error by 5.93 times, and PSNR prediction error by 1.20 times, all highly significant ($ p<0.0001$). By decreasing validation loss on unseen augmentations, NovisVQ learned to approximate the underlying full-reference metrics rather than memorizing distortion-specific artifacts, demonstrating the necessity of temporal context for video quality assessment.

\begin{figure}[H] 
    \centering
    \includegraphics[width=\linewidth]{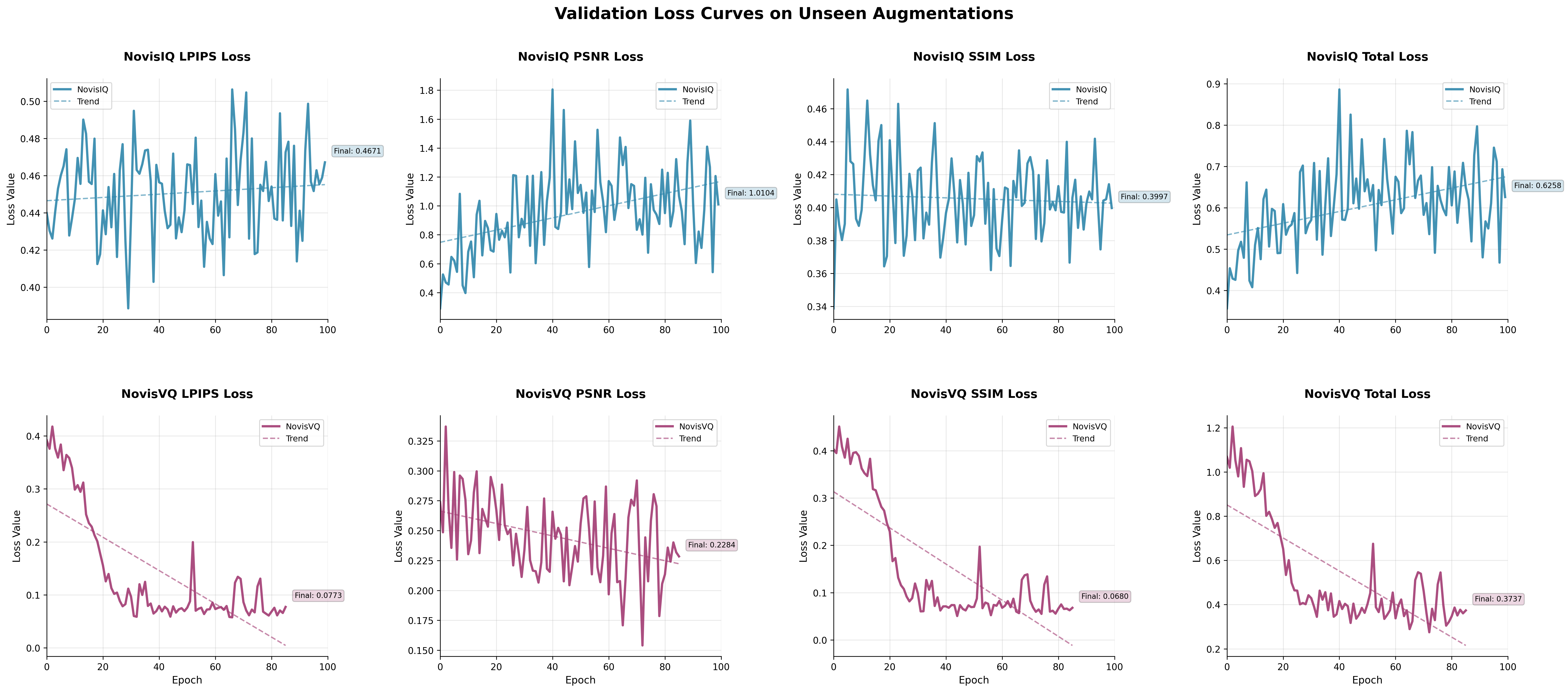}
    \caption{Validation loss curves for NovisIQ and NovisVQ across LPIPS, PSNR, SSIM, and total loss. The video-based model consistently converges, while the image-based model diverges, highlighting the necessity of streaming models for frame quality evaluation in video object detection systems.}
    \label{fig:val_curves}
\end{figure}

\subsection{Transfer to Real World Data}

We tested on the GOPRO\_Large dataset \cite{Nah_2017_CVPR}, which contains motion-blurred GoPro videos with sharp ground truth frames. This presents a significant generalization challenge: our models were trained exclusively on synthetic degradations (Gaussian blur, brightness, JPEG compression), while GOPRO contains realistic motion blur generated by averaging consecutive high-speed frames to simulate camera exposure \cite{Nah_2017_CVPR}. This temporal integration approach produces motion blur from actual camera and scene movement, representing a fundamentally different degradation mechanism than our training augmentations and posing a challenging real-world transfer problem in computer vision which can lead to false negatives and broken object tracks.

\begin{figure}[H] 
    \centering
    \includegraphics[width=\linewidth]{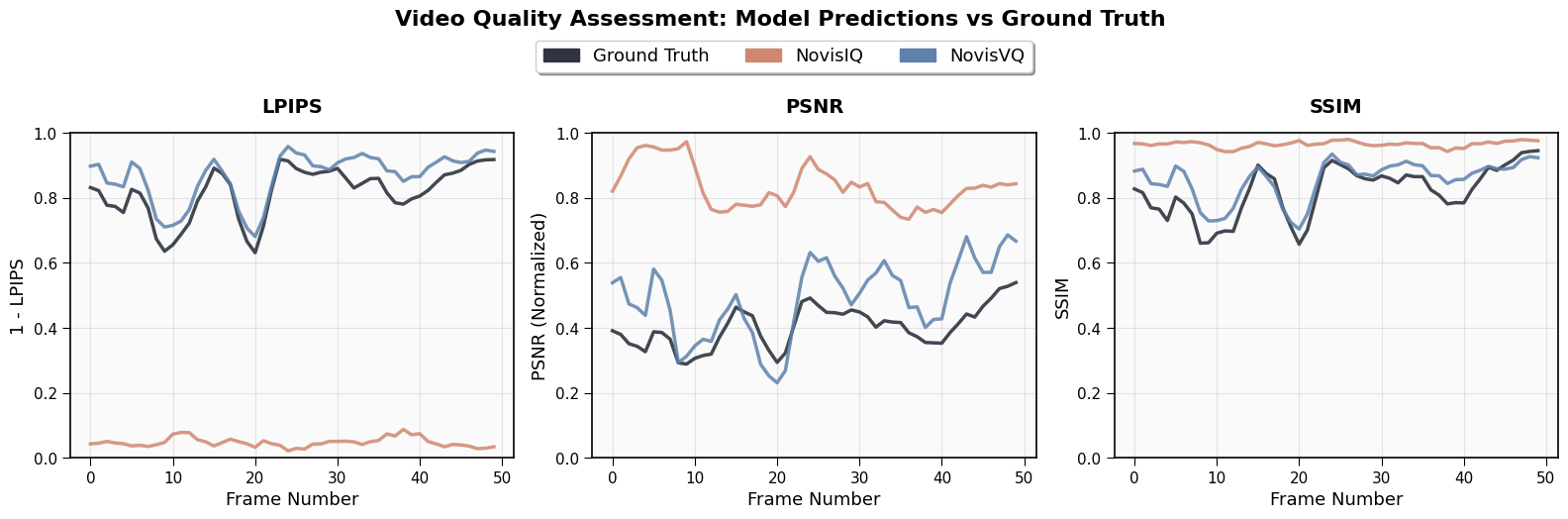}
    \caption{Quality metric predictions on real-world motion blur. NovisVQ closely tracks ground truth across all metrics, while NovisIQ shows poor correlation, demonstrating temporal context enables better generalization from synthetic to real-world degradations.}
    \label{fig:real_world}
\end{figure}

We evaluated NovisVQ against our image-based baseline, NovisIQ, across 1,111 frames from 11 real-world motion-blurred videos not seen during training. Despite the synthetic-to-real degradation shift, NovisVQ demonstrates substantial improvements from NovisIQ: 84.7\% MAE reduction for LPIPS (0.670 → 0.102), 65.4\% for PSNR (0.394 → 0.136), and 31.8\% for SSIM (0.156 → 0.107). Correlations with ground truth further confirm NovisVQ's superiority (LPIPS: r=0.66, PSNR: r=0.55, SSIM: r=0.57) versus NovisIQ's near-zero or negative correlations. 

For comparison, we also computed correlations with BRISQUE, a no-reference metric trained on human opinion scores rather than full-reference metrics, which achieved weaker correlations (LPIPS: r=0.27, PSNR: r=0.24, SSIM: r=0.21). NovisVQ's correlation with full-reference metrics positions it as a valuable addition to the no-reference VQA world, offering objective quality estimation without requiring pristine reference frames or training on subjective human labels. By outperforming BRISQUE in correlation with full-reference metrics, NovisVQ demonstrates that temporal, opinion-unaware modeling can surpass traditional image-based approaches in predicting objective quality measures. The visualization shows a representative 50-frame segment demonstrating NovisVQ predictions closely tracks each ground truth full-reference metric.

\begin{table}[h]
\centering
\caption{Generalization to Real-World Motion Blur: MAE and Correlation Analysis on GOPRO Dataset}
\label{tab:gopro_results}
\begin{tabular}{lccccccc}
\toprule
\multirow{2}{*}{\textbf{Model}} & \multicolumn{3}{c}{\textbf{MAE}} & & \multicolumn{3}{c}{\textbf{Correlation (r)}} \\
\cmidrule{2-4} \cmidrule{6-8}
& \textbf{LPIPS} & \textbf{PSNR} & \textbf{SSIM} & & \textbf{LPIPS} & \textbf{PSNR} & \textbf{SSIM} \\
\midrule
NovisIQ  & 0.670 & 0.394 & 0.156 & & 0.03 & -0.30 & -0.08 \\
NovisVQ  & \textbf{0.102} & \textbf{0.136} & \textbf{0.107} & & \textbf{0.66} & \textbf{0.55} & \textbf{0.57} \\
BRISQUE  & --- & --- & --- & & 0.27 & 0.24 & 0.21 \\
\bottomrule
\end{tabular}
\end{table}

\section{Conclusion}
\label{sec:Conclusion}

We presented NovisVQ, a streaming, no-reference, opinion-unaware video quality assessment model that predicts full-reference metrics (LPIPS, PSNR, SSIM) directly from degraded video without requiring pristine references or training on human opinion labels. Through comparison against our image-based baseline NovisIQ, we demonstrated that temporal context is essential for robust frame quality prediction.

Our key findings show streaming architectures substantially outperform frame-independent approaches. While NovisIQ overfit to training augmentations (validation loss increased 81.98\%), NovisVQ generalized strongly with 65.02\% validation loss reduction. On real-world GOPRO motion blur, NovisVQ achieved 84.7\% MAE reduction and strong correlations (r=0.55-0.66), significantly outperforming NovisIQ and BRISQUE (r=0.21-0.27).

These results validate that temporal modeling enables learning generalizable perceptual quality representations rather than memorizing artifacts. By training on synthetic degradations yet generalizing to real-world motion blur, NovisVQ demonstrates the viability of scalable, self-supervised VQA without reference videos or human annotations. Future work should explore direct downstream task prediction, alternative temporal architectures, and broader degradation types for quality-aware video processing in autonomous systems and real-time applications.

\section*{Acknowledgements}
This work was performed under the auspices of the U.S. Department of Energy by Lawrence Livermore National Laboratory under Contract DE-AC52-07NA27344 with funding from the lab and the U.S. Navy. Release number: LLNL-JRNL-2013272

\bibliographystyle{unsrt}
\bibliography{sample}

\appendix

\section{Model Architecture}

\subsection{Multiscale ResNet Image Encoder} 
The visual encoder is a compact modification of the standard ResNet-18 \cite{he2015deep}, designed to extract multiscale spatial features from video frames. The encoder takes a single frame as input and outputs hierarchical feature maps at different resolutions:

Input Processing: The initial convolution layer reduces the input image dimensions by a factor of 2 using a 7×7 kernel with stride 2, followed by batch normalization, ReLU activation, and a 3×3 max-pooling layer.

Residual Layers: We use a three-layer ResNet variant with BasicBlock modules and reduced channel widths:
\begin{itemize}
\item Layer1: 2 residual blocks, 32 channels

\item Layer2: 2 residual blocks, 64 channels (stride = 2)

\item Layer3: 2 residual blocks, 128 channels (stride = 2)
\end{itemize}

This results in four outputs per frame:
$x_1$ (post-initial conv), $x_2$, $x_3$, and $x_4$ corresponding to increasingly abstract spatial representations.

\subsection{Temporal Feature Modeling with LSTMs}
To capture temporal coherence and motion-related degradations, we apply separate lightweight LSTM networks to the encoder outputs ($x_1$ to $x_3$). The LSTM units are single-layer, unidirectional, and operate across time for each spatial feature map:
\begin{itemize}
\item $x_1$ and $x_2$: processed using LSTMs with 32 hidden units.

\item $x_3$: processed with an LSTM with 64 hidden units.

\end{itemize}

Before feeding into the LSTMs, spatial feature maps are sub-sampled by a factor of 2 (to preserve local structure while reducing size) and grouped into smaller spatial patches to mitigate memory costs. Each group is treated as a sequence and passed through the LSTM, and the outputs are later recombined to reconstruct full feature maps across time.

For chunked inference, LSTM hidden states are propagated across overlapping video chunks, allowing temporal continuity across segment boundaries.

\subsection{Quality Prediction MLP}
A lightweight multi-layer perceptron (MLP) aggregates temporal features and outputs the predicted quality metrics:

Inputs: The outputs of the LSTM-processed features ($h_1$, $h_2$, $h_3$) and the final ResNet feature map ($x_4$).

Averages: Global average pooling is applied over spatial dimensions.

Concatenation: All pooled features are concatenated into a single vector of 512 dimensions.

MLP Layers:
\begin{itemize}
\item Linear(512 → 256), ReLU, Dropout(0.1)

\item Linear(256 → 128), ReLU, Dropout(0.1)

\item Linear(128 → 3), followed by a Sigmoid activation

\end{itemize}

The final output is a tensor of shape $(b, t, 3)$, corresponding to frame-wise predictions for LPIPS, PSNR, and SSIM.

\begin{figure}[h]
\centering
\includegraphics[width=\textwidth]{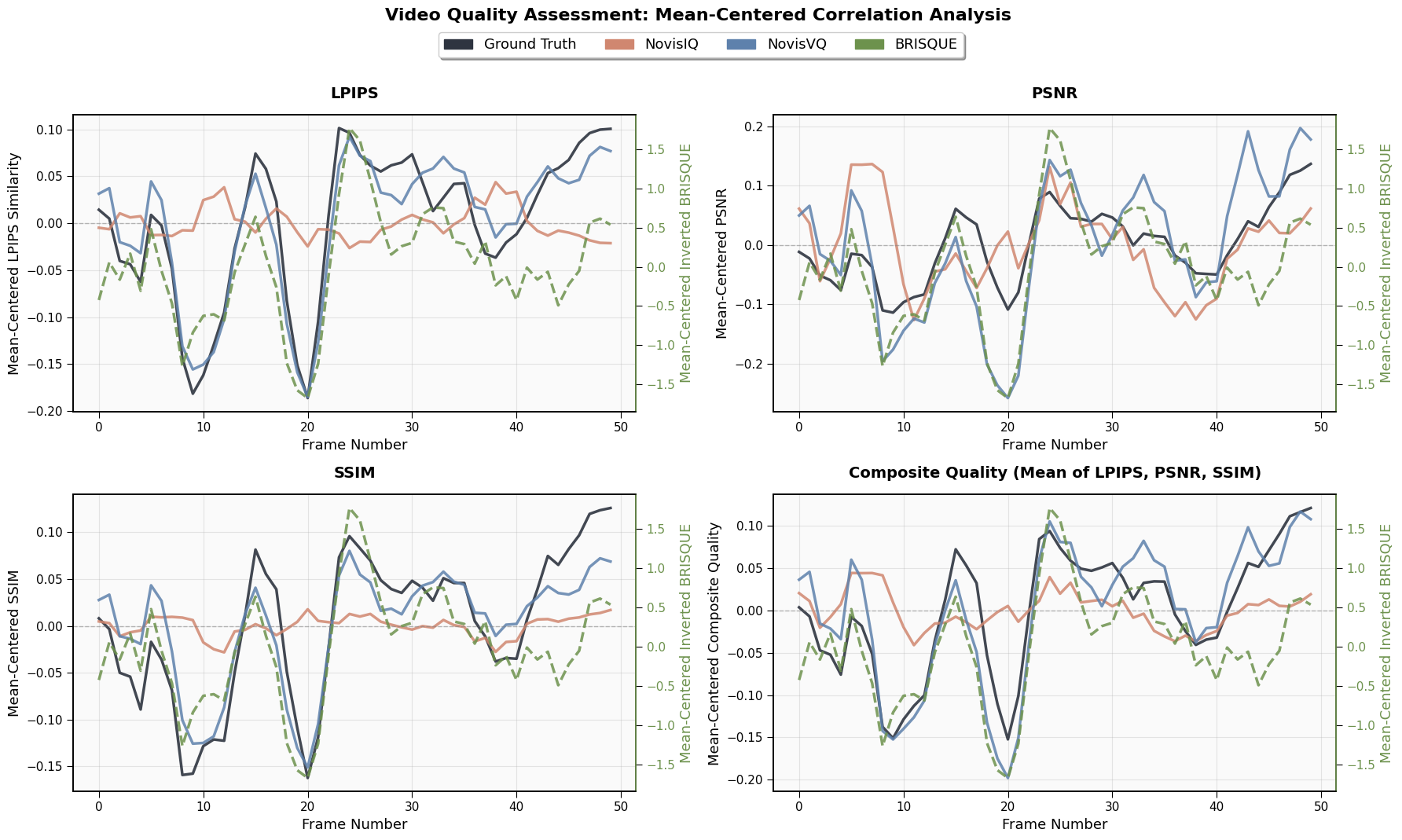}
\caption{Mean-centered correlation analysis comparing NovisVQ, NovisIQ, and BRISQUE predictions against ground truth full-reference metrics on real-world motion-blurred video. Each subplot shows mean-centered predictions for individual metrics (LPIPS, PSNR, SSIM) and a composite quality score (mean of all three metrics). NovisVQ (blue) demonstrates substantially stronger correlation with ground truth (black) compared to both NovisIQ (orange) and BRISQUE (green, shown on secondary axis). The composite metric plot provides an apples-to-apples comparison between our video-based approach and the image-based BRISQUE baseline, highlighting the value of temporal context for quality assessment. All metrics are smoothed with a 3-frame moving average and mean-centered for visualization. Correlation coefficients (r) are displayed in legends.}
\label{fig:appendix_correlation}
\end{figure}

\end{document}